\documentclass[runningheads]{llncs}

\usepackage{accv}

\usepackage{accvabbrv}

\usepackage{graphicx}
\usepackage{booktabs}
\usepackage[T1]{fontenc}
\usepackage{multirow}
\usepackage{amssymb}
\usepackage{pifont}
\usepackage{units}

\usepackage[accsupp]{axessibility}  

\usepackage{hyperref}

\usepackage{orcidlink}

\begin{document}

\title{Boosting Few-Shot Detection with Large Language Models and Layout-to-Image Synthesis} 

\titlerunning{Boosting Few-Shot Detection}

\author{Ahmed Abdullah\textsuperscript{1} \and
Nikolas Ebert\textsuperscript{1}\and
Oliver Wasenm\"uller\textsuperscript{1}}

\authorrunning{Abdullah et al.}

\institute{\textsuperscript{1}Mannheim University of Applied Sciences, Germany\\
\email{\{a.abdullah, n.ebert, o.wasenmueller\}@hs-mannheim.de}
}

\maketitle
\begin{abstract}
  Recent advancements in diffusion models have enabled a wide range of works exploiting their ability to generate high-volume, high-quality data for use in various downstream tasks. One subclass of such models, dubbed Layout-to-Image Synthesis (LIS), learns to generate images conditioned on a spatial layout (bounding boxes, masks, poses, etc.) and has shown a promising ability to generate realistic images, albeit with limited layout-adherence. Moreover, the question of how to effectively transfer those models for scalable augmentation of few-shot detection data remains unanswered. Thus, we propose a collaborative framework employing a Large Language Model (LLM) and an LIS model for enhancing few-shot detection beyond state-of-the-art generative augmentation approaches. We leverage LLM’s reasoning ability to extrapolate the spatial prior of the annotation space by generating new bounding boxes given only a few example annotations. Additionally, we introduce our novel layout-aware CLIP score for sample ranking, enabling tight coupling between generated layouts and images. Significant improvements on COCO few-shot benchmarks are observed. With our approach, a YOLOX-S baseline is boosted by more than 140\%, 50\%, 35\% in mAP on the COCO 5-,10-, and 30-shot settings, respectively.
\end{abstract}

\section{Introduction}
\label{sec:intro}

Object detection and image classification models have enjoyed significant improvements over the years, thanks to both architectural improvements \cite{vitscaling, ren2015faster, zhang2022dino, liu2022convnet, ebert2023plg, ebert2023transformer} and extensively annotated datasets \cite{deng2009imagenet, lin2014microsoft, zhou2017scene, cruz2020sviro}. The advantages that a high volume, appropriately annotated dataset can bring are evident in the rise of performance, applicability, and robustness of state-of-the-art computer vision models across many domains and tasks. However, such datasets are difficult to collect in a transparent fashion \cite{scheuerman2021datasets}, and even more difficult to annotate, often requiring extensive manual effort to provide accurate labelling and annotating. 

This bottleneck in scaling datasets becomes a limitation that simply training larger models may not be able to overcome. 
Additionally, the distributions of established datasets may not adequately reflect the full range of real-world concepts (e.g. specific species of animals). This lack of coverage can result in suboptimal downstream performance of models deployed in scenarios where such concepts are occurring more frequently.
\begin{figure}[tb]
  \centering
  \includegraphics[width=\textwidth]{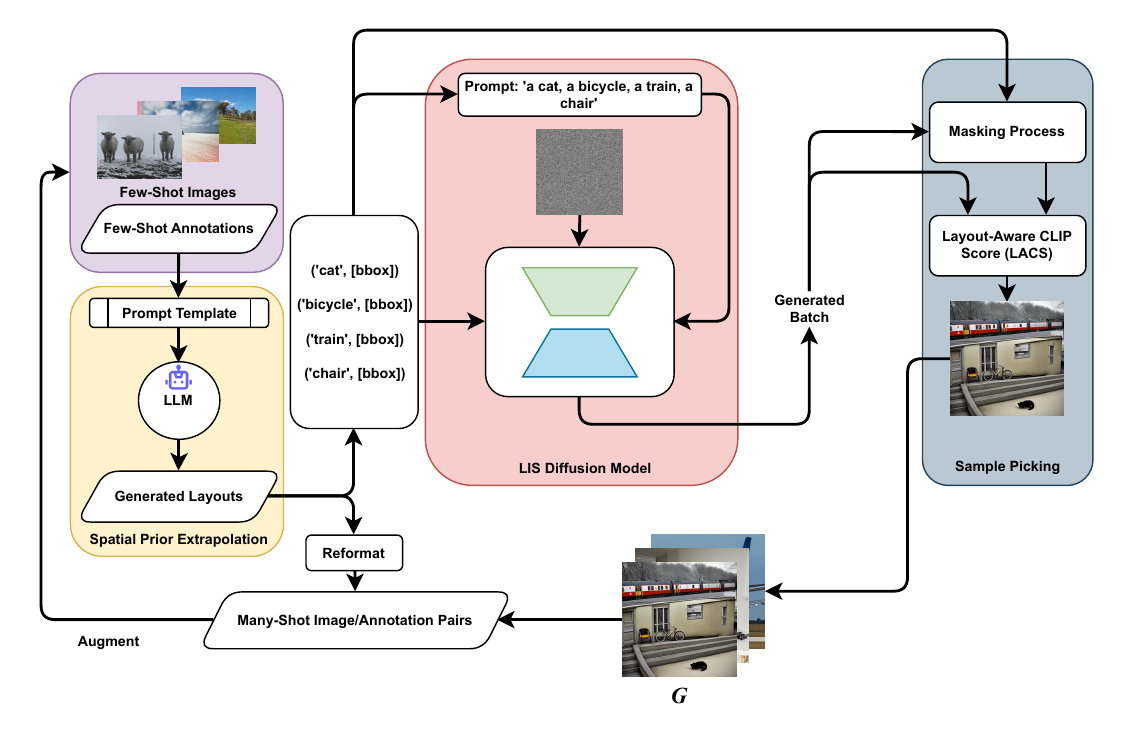}
  \caption{Overview of proposed framework. First, we employ a pretrained LLM to generate new layouts with the help of a prompt template and the available few-shot annotations. Next, we generate new images using a layout-to-image synthesis diffusion model, conditioned on the newly created layouts. Finally, we employ a layout-aware CLIP score (LACS) to rate the generated samples, and construct a generated set of images \textit{\textbf{G}} with high layout-adherence by picking the highest scoring images from a given batch. After reformatting the generated layouts into detection annotations, the resulting image-annotation pairs are used to augment the few-shot detection data.
  }
  \label{fig:method_overview}
\end{figure}

Thus, few-shot learning has seen a noticeable emergence in research \cite{brown2020language, zhou2023thin, zhou2021flipda} concerned with training models where only a few samples are available, requiring approaches that can generalise to rarely seen concepts.
One common approach to account for low-frequency, novel objects is by employing data augmentation pipelines. These range from basic image manipulation operations such as rotation, flipping, scaling, or cropping, to more advanced combinational augmentation strategies, such as CutMix \cite{yun2019cutmix}, MixUp \cite{zhang2017mixup}, AugMix \cite{hendrycks2019augmix} and Mosaic \cite{ge2021yolox}.

Recently, generative augmentation has emerged as a result of the advancement of image generation methodologies \cite{rombach2022high, song2020denoising, karras2019style, karras2020training}. A plethora of works research the capability of such models to augment existing datasets \cite{oehrisynth, bansal2023leaving, fang2024data, wu2023datasetdm, reichardt2024text3daug}. 
In the context of image classification, augmentation-by-generation \cite{zhou2023thin, bansal2023leaving} is a straightforward process, provided that the generated sample is free of artifacts. This is primarily attributable to the fact that the spatial location of the concept within the synthesized image is not a major concern in this process.
However, generative augmentation for detection and segmentation requires robust conditional generators that can synthesize instances of objects strictly within the conditional spatial boundaries, without compromising the visual quality of the underlying samples.

Hence, Layout-to-Image Synthesis models (LIS) \cite{zhang2023adding, mou2024t2i, xue2023freestyle} have emerged, giving rise to generative augmentation methods for detection and segmentation tasks \cite{fang2024data, yang2024freemask, jia2023dginstyle, wu2023datasetdm} that employ diffusion models in their pipelines to provide a performance boost to the downstream detectors and segmentors. Nevertheless, many generative augmentation methods are either not applicable when taking the few-shot detection scenario into account \cite{yang2024freemask, jia2023dginstyle}, fail to scale on higher augmentation ratios \cite{fang2024data}, or utilize pseudo-labelling in the pipeline instead of an LIS model \cite{feng2024instagen, wu2023datasetdm}. The latter approach leads to subpar image-annotation alignment due to the limitations of pseudo-labelling.

We hypothesize that to effectively reach high quality generative dataset augmentation for few-shot detection tasks, the spatial prior in the annotation space must be expanded in an automated manner, without additional manual effort. This approach ensures that not only the appearance of objects is augmented, but also their location within the scene. Inspired by the progress achieved by Large Language Models (LLM), we employ an LLM-based approach for generating new layouts, effectively extrapolating the spatial prior of the few-shot annotation space. We combine the previous extrapolation step with an LIS model to unlock scalable, high quality generative augmentation for few-shot detection.

In addition, we discover that only a few LIS approaches can generate realistic images while maintaining layout conformity. 
To ensure close image-layout alignment and realistic objects, we propose our novel Layout-Aware CLIP Score (LACS) for ranking generated images based on both realism and layout-adherence, effectively eliminating noisy samples. CLIP score has been used in prior works to rate the realism of generated images and for calculating the alignment between images and their respective text captions \cite{radford2021learning, hessel2021clipscore, wang2024instancediffusion}. However, we aim to employ a more advanced scoring scheme to account for hallucinations outside of the conditional regions.

In summary, we propose our novel collaborative LLM-LIS generative augmentation framework (shown in Figure \ref{fig:method_overview}) for few-shot object detection that scales well beyond state-of-the-art approaches. We utilize an LLM for spatial prior extrapolation given only a small number of ground truth layouts. Furthermore, we employ our layout-aware CLIP score to enable close image-layout alignment for superior downstream performance. We extensively evaluate our approach on the COCO \cite{lin2014microsoft} few-shot detection benchmark and achieve substantial improvements in mean average precision (mAP) on the 5-, 10-, and 30-shot settings.

\section{Related Work}
\label{sec:rw}

\subsection{Few-Shot Object Detection}
To tackle the challenge of data scarcity and the need to detect novel categories in the wild, few-shot detection aims to train a detector model using only a small number of instances of those categories in the training data. While many paradigms have been proposed on how to train a detector for few-shot tasks \cite{wang2019meta, wang2020frustratingly}, the most common way to handle few-shot detection regimes is pretrain-into-finetune \cite{wang2020frustratingly}. This is accomplished by pretraining the detection model on high-frequency base categories with ample annotations, and then finetuning on the low-frequency, novel categories of objects while the backbone of the detector model remains frozen.

\subsection{Layout-to-Image Synthesis}
Regarded as the inverse task of object detection, Layout-to-Image Synthesis (LIS) aims to generate images by providing the model with layout information typically in the form of bounding boxes, masks, edge information, or scribbles. Earlier works \cite{li2020bachgan, sun2019image, tang2020edge} utilize Generative Adversarial Networks (GAN) \cite{goodfellow2020generative} to synthesize images conditioned on a specified semantic or bounding box layout.  
Inspired by recent advancements in Text-to-Image (T2I) diffusion models \cite{rombach2022high, luo2023latent}, many works \cite{zhang2023adding, xue2023freestyle, wang2024instancediffusion, xie2023boxdiff} have emerged to introduce granular location control of concepts synthesized by the diffusion model. This can be done either in a training-free, or a training-based manner.

\textbf{Training-Free LIS}
To bypass the computational cost of training auxiliary modules and maintain a wide concept coverage, training-free LIS methods \cite{xie2023boxdiff, chen2024training, mo2024freecontrol} propose leveraging a frozen latent diffusion model \cite{rombach2022high} for spatially-controllable T2I generation. This is accomplished by guiding the diffusion model features towards the layout condition regions during inference. However, most training-free LIS approaches result in rough layout-adherence, rendering them of limited applicability for few-shot detection.

\textbf{Training-Based LIS}
Contrary to training-free LIS, training-based methods \cite{zhang2023adding, xue2023freestyle, mou2024t2i, li2024aldm, lv2024place, wang2024instancediffusion, li2023gligen} integrate auxiliary modules with the frozen encoder-decoder blocks of the diffusion model to learn additional control features by training those modules on detection and segmentation datasets \cite{zhou2017scene, lin2014microsoft}. Gligen \cite{li2023gligen} uses bounding boxes to formulate grounding tokens, employing an additional gated attention layer to achieve LIS. The recently proposed InstanceDiffusion \cite{wang2024instancediffusion} enables fine-grained instance-level control with flexible layout definitions by tokenizing the location information per-instance and fusing the resulting features with the frozen diffusion model backbone, achieving impressive LIS performance.

\subsection{Applications of Generative Augmentation for Object Detection}

Fang et al. \cite{fang2024data} propose a framework utilizing visual priors (e.g. Holistic Edge \cite{xie2015holistically}, semantic segmentation maps) to augment few-shot detection by generating images with ControlNet \cite{zhang2023adding} using the acquired visual priors. InstaGen and DatasetDM \cite{wu2023datasetdm, feng2024instagen} incorporate a trainable detector to pseudo-label images generated with a latent diffusion model \cite{rombach2022high} by exploiting rich attention maps from the diffusion model. Lin et al. \cite{lin2023explore} implement a copy-paste pipeline to enhance few-shot object detection.

While both Fang et al. \cite{fang2024data} and InstaGen \cite{feng2024instagen} offer promising results in object detection performance, they fall short due to either non-scalability with higher augmentation ratios \cite{fang2024data} or insufficient image-annotation alignment of generated data due to noisy pseudo-labelling \cite{feng2024instagen}. Our approach seeks to solve those shortcomings by extrapolating the spatial prior of bounding box distributions to achieve scalable, high quality generative augmentation. Instead of relying on post-generation pseudo-labelling, we deploy a pretrained LIS diffusion model coupled with our novel Layout-Aware CLIP Score (LACS) to ensure close image-layout alignment.

\section{Method}
\label{sec:method}

In this section, we describe the key modules of our generative augmentation framework. As seen in Figure \ref{fig:method_overview}, we employ three modules to obtain high quality generated data. First, an LLM-based module extrapolates the spatial prior of the few-shot annotation data by generating bounding box layouts. The generated layouts are then used to guide an LIS model for image synthesis. This is followed by our novel layout-aware CLIP score, which is used to rate the generated batch of images for sample picking.

\subsection{LLM-based Spatial Prior Extrapolation}
\label{subsec:spe}
\begin{figure}[t]
  \centering
  \includegraphics[width=\textwidth]{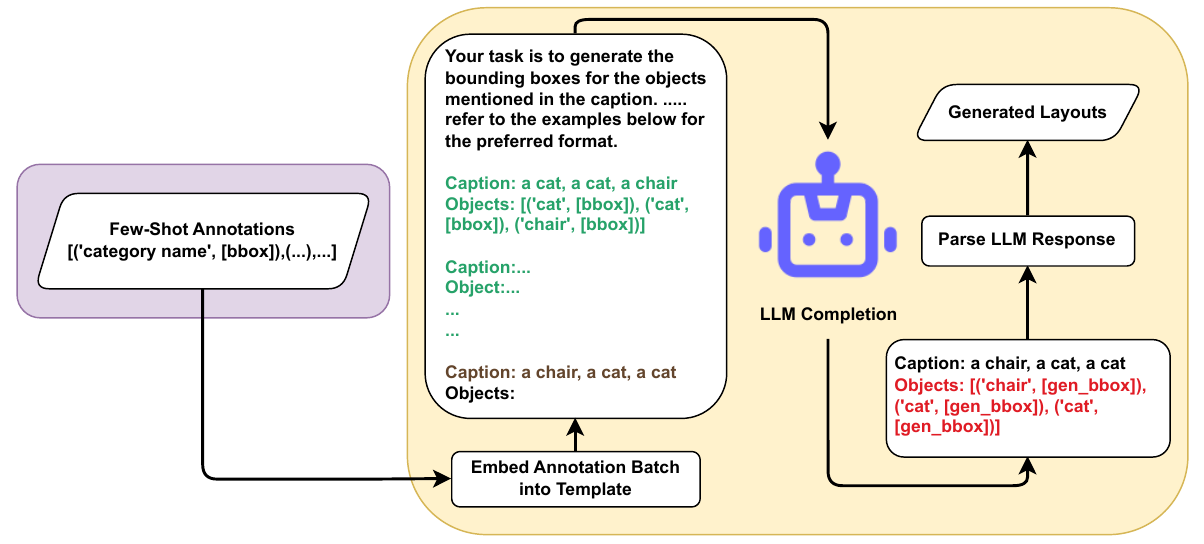}
  \caption{LLM-based spatial prior extrapolation. First, we embed a batch of few-shot annotations into a prompt template by formatting them as layout descriptions. The embedded layout descriptions serve as in-context examples (in green) to steer the text generation process. We then prompt the LLM to complete a caption of randomly reordered objects (brown) from one of the layout descriptions and obtain a response containing generated bounding boxes (red). Finally, we parse the response to obtain new layouts.
  }
  \label{fig:llm_spe}
\end{figure}

Motivated by the emerging application of language models as layout generators \cite{xie2023visorgpt, lian2023llm, feng2024layoutgpt}, we incorporate a pretrained large language model for spatial prior extrapolation (SPE) via generation of novel layouts.

To elaborate, the spatial prior encompasses the existing layout annotations in the few-shot data, and is presented to the generator model (in our case, the LLM) as in-context examples. The task of the generator model is to \textit{extrapolate} from the presented examples, and generate plausible layouts to be used for image generation via LIS, hence the term \textit{Spatial Prior Extrapolation}.

We employ LLM auto-completion with a pretrained Mixtral-8x7B-Instruct-v0.1. \cite{jiang2024mixtral}. We choose Mixtral as our LLM due to its' competitive performance in layout generation when compared to closed-source models such as GPT-3.5 and GPT-4 in spatial and numeracy reasoning tasks, as tested by LMD\cite{lian2023llm}. Similar to LMD \cite{lian2023llm}, we first compose a prompt template. The template consists of detailed bounding box layout generation instructions for the LLM to follow, in addition to spatial restrictions with respect to box placement and image dimensions. We then embed a batch of per-image ground truth annotations  \( A_{b} \) as context examples into the prompt template by formatting them as layout descriptions (e.g. caption:'a cat, a car, a person', objects: ['cat',[bbox1], 'car', [bbox2], 'person', [bbox3]]), as seen in Figure \ref{fig:llm_spe}.

We process the batch \( A_{b} \) by concatenating the captions of the ground truth layout descriptions (one caption per auto-completion) after randomizing the occurrence-order of the objects within the captions. The LLM is then used to auto-complete the prompt, generating object bounding boxes corresponding to the concatenated caption. To induce generation diversity, we also randomize the order of examples in the prompt template for each auto-completion.

We repeat the auto-completion for \textbf{$\alpha$} times per batch element, where \textbf{$\alpha$} is the augmentation ratio (number of synthetic layouts over real layouts). The bounding box layouts generated by the LLM are subsequently parsed and utilized in the LIS process and as annotations for detector training.

\subsection{Layout-to-Image Synthesis Diffusion Model}
\label{subsec:instdiff}

At the core of our framework, we employ a pretrained LIS diffusion model for image generation. 
In contrast to typical text-to-image latent diffusion models \cite{rombach2022high}, an LIS diffusion model accepts spatial layouts as conditional signals to guide the results of text-to-image generation. 
\newline
For generating images from a layout and a text condition, a batch of Gaussian noise latent codes \(Z \in \mathbb{R}^{b \times h \times w \times c}\) is initialized,
where \( b \), \( h \), \( w \), \( c \) correspond to the batch size, height, width, and number of channels of the latent code, respectively. The latent denoising network \textbf{LDN} of the LIS model produces the denoised latent codes \(\Tilde{Z} \) as

\begin{equation}
\label{eq:lis1}
\centering
\Tilde{Z} = \textbf{LDN}(Z, p, l, t),
\end{equation}

where \( p \) corresponds to the text prompt, \( l \) the spatial layout condition, and \( t \) the number of denoising time steps.
\newline
The denoised latents \(\Tilde{Z} \) are then decoded by the latent decoder $\mathcal{D}$ to generate a batch of images \( \Tilde{X} \) as

\begin{equation}
\label{eq:lis2}
\centering
\Tilde{X} = \mathcal{D}( \Tilde{Z} ).
\end{equation}

In Section \ref{sec:lis_comp}, we ablate over different pretrained State-of-the-Art LIS models with both masks and bounding boxes as conditional layouts. We discover that InstanceDiffusion \cite{wang2024instancediffusion} with bounding box conditions offers the best synthesis in terms of layout-adherence and overall image quality.

\subsection{Layout-Aware CLIP Score}
\label{subsec:clipscore}
\begin{figure}[tb]
  \centering
  \includegraphics[width=\textwidth]{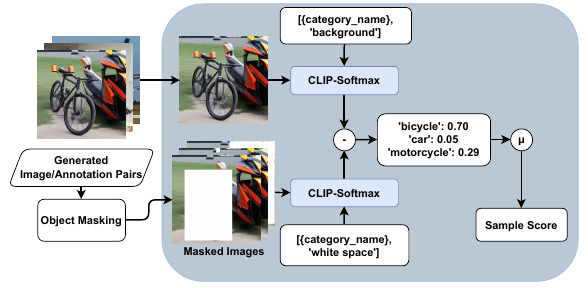}
  \caption{Overview of LACS. First, we create  \( \mathbf{n} \) masked images from a generated image, where  \( \mathbf{n} \) is the number of object categories in the image. Next, for each category, we perform zero-shot classification on both the generated image and the masked image and obtain a per-category layout-adherence score by subtracting the two classification scores. We average over all categories to arrive at the final sample score.
  }
  \label{fig:layout_aware_clip_score}
\end{figure}

To ensure a close image-layout alignment, we integrate a novel layout-aware CLIP score into our framework (shown in Figure \ref{fig:layout_aware_clip_score}). For each image \( \Tilde{x} \) in a batch of LIS generated images \( \Tilde{X} \), we first obtain the CLIP \cite{radford2021learning} softmax score \( CS \) for each category present in the layout

\begin{equation}
\label{eq:cs1}
\centering
CS(text_{n}, \Tilde{x}) = S_{0}(C_{logits}(\Tilde{x}, \{text_{n},'background'\})),
\end{equation}
where \( text_{n} \) corresponds to the text of the \( n \)'th category in the layout, \( C_{logits} \) the CLIP cosine similarity logits, and \( S_{0}\) denotes the softmax output of the first logit from the CLIP inference.

Next, we produce a masked image \( \Tilde{x}_{masked_n} \) from the generated image by masking the bounding boxes of the respective category with white pixels. We obtain the masked CLIP softmax score \( CS_{mask} \) by inferring on the masked image and switching the second text prompt from \( 'background' \) to \( 'white space' \) as
\begin{equation}
\label{eq:cs2}
\centering
CS_{mask}(text_{n}, \Tilde{x}_{masked_n}) = S_{0}(C_{logits}(\Tilde{x}_{masked_n}, \{text_{n},'white space'\})).
\end{equation}

The text prompt \( 'background' \) is introduced as a placeholder to quantify the presence of the object category in the image for the first score \( CS \). In the masked score \( CS_{mask} \), the text prompt \( 'white space' \) is used for inducing a lower value in the logit corresponding to the object category text \( text_{n} \), when a generated image \( \Tilde{x} \) contains little to no out-of-layout hallucinations. The opposite occurs when said hallucinations are present in the generated image, leading to a higher logit value for \( text_{n} \).
Finally, we obtain the final Layout-Aware CLIP Score (LACS) for image \( \Tilde{x} \) by subtracting \( CS_{mask} \) from \( CS \) for all \( n \) categories and calculating the mean score as
\begin{equation}
\label{eq:cs3}
\centering
LACS(\Tilde{x}) = \frac{1}{n} \sum_{1}^{n} CS(text_{n},\Tilde{x}) - CS_{mask}(text_{n},\Tilde{x}_{masked_n}).
\end{equation}

\begin{figure}[tb]
  \centering
  \includegraphics[width=0.9\textwidth]{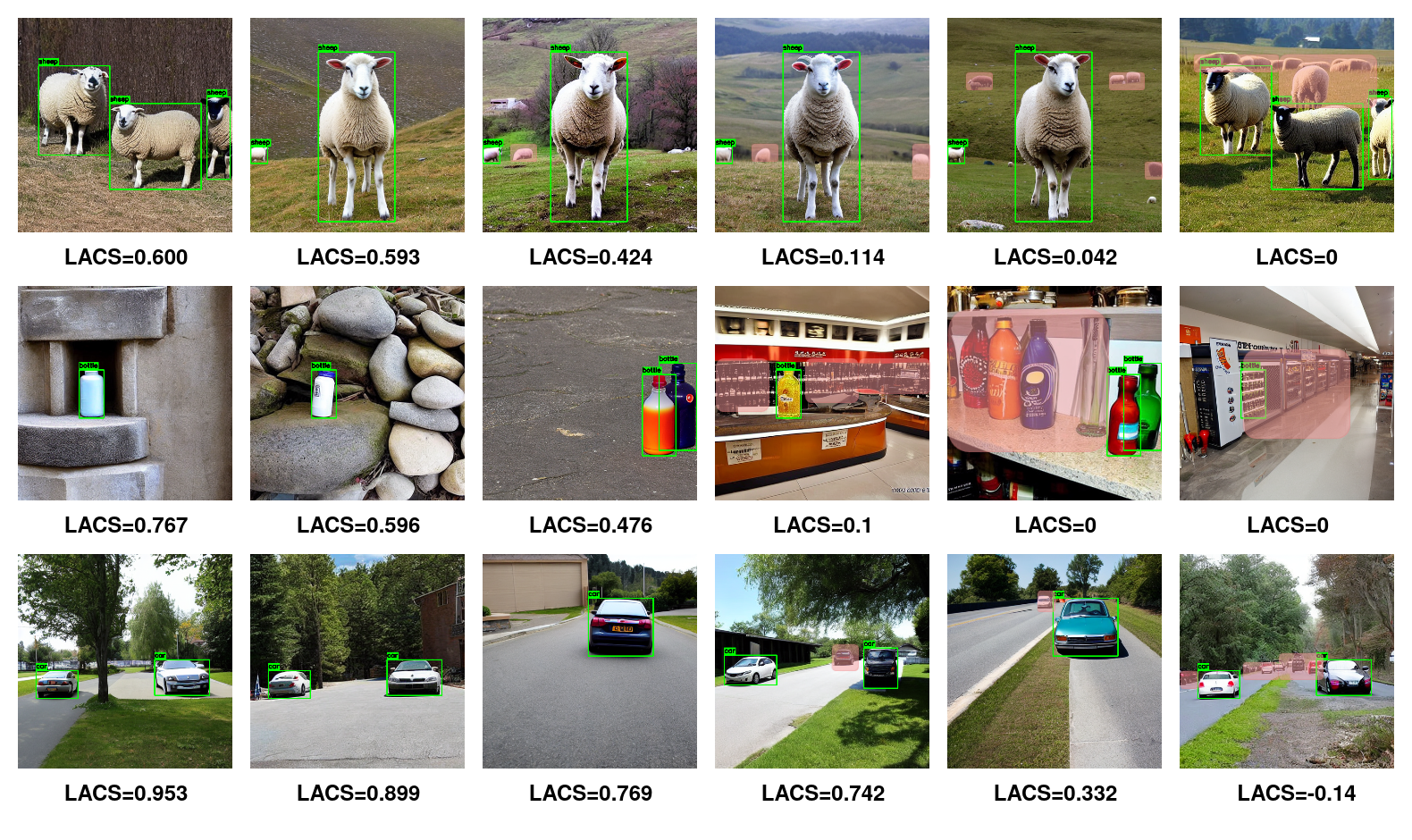}
  \caption{Samples generated with InstanceDiffusion \cite{wang2024instancediffusion} and their LACS score. Green boxes highlight the conditional layout, while red regions highlight out-of-layout hallucinations.
  }
  \label{fig:lacs_gen}
\end{figure}

The score above is used to sort the individual images \( \Tilde{x} \) in the generated batch \( \Tilde{X} \), where the best ranking images contain the least out-of-layout hallucinations, and vice-versa. We utilize the score for sample picking in our experiments (LACS-SP), and for calculating the mean layout-adherence of generated images (mLACS) for quality comparisons. Figure \ref{fig:lacs_gen} shows the LACS score for images generated with InstanceDiffusion \cite{wang2024instancediffusion} and its' correlation with out-of-layout hallucinations. 

\section{Experiments}
\label{sec:exp}

\subsection{Experiment Setup}
For the evaluation of our method, we conduct our experiments using the COCO \cite{lin2014microsoft} detection dataset. We evaluate under two settings: the established few-shot novel category finetuning \cite{kang2019few, wang2020frustratingly, fang2024data}, and all-category training using a small synthetic dataset. 
Unless otherwise stated, we use Mixtral-8x7B-Instruct-v0.1 \cite{jiang2024mixtral} for bounding box layout generation and embed its prompt template (outlined in Section \ref{subsec:spe}) with batches \( A_{b} \) of five example layouts from the ground truth annotations. Here, we set the augmentation ratio $\alpha$ to 4. When not utilizing spatial prior extrapolation with Mixtral, we simply oversample the ground truth annotations with x-axis flipping to reach the desired augmentation ratio. For layout-to-image synthesis, we utilize a pretrained InstanceDiffusion \cite{wang2024instancediffusion} with bounding boxes as layouts and set the number of denoising steps \( t \) to 50, the guidance scale to 7.5, grounding-alpha (percentage of timesteps using grounding inputs) to 0.8, and MIS (percentage of timesteps using the multi-instance sampler) to 0.36. We generate a batch \( \Tilde{X} \) of five images, rank them with our novel LACS score, and pick the top-1 image from the batch. Unless otherwise stated, we augment the few-shot real data with the resulting generated data. All detectors are set to a fixed seed for reproducibility when finetuning.

\subsection{Evaluation on COCO Detection}

\subsubsection{Few-Shot Novel Category Finetuning}

\begin{table}[t]
\scriptsize
\centering
\caption{Main results over novel COCO \cite{lin2014microsoft} 5-,10-, and 30-shot settings. We compare our approach against state-of-the-art generative augmentation methods \cite{fang2024data, feng2024instagen} in few-shot detection settings and showcase the effect of spatial prior extrapolation (SPE) and top-1 sample picking w.r.t. our layout-aware CLIP Score (LACS-SP). We utilize InstanceDiffusion \cite{wang2024instancediffusion} in our framework and compare downstream performance with and without our contributions (SPE and LACS-SP). Results for Fang et al. \cite{fang2024data} correspond to their best augmentation ratio $\alpha$ of 1 as reported in their findings. For comparison with InstaGen \cite{feng2024instagen}, we synthesize with an augmentation ratio $\alpha$ of 4 with both their method and ours.}
\begin{tabular}{llcccccccc}
\hline
\multicolumn{2}{l}{\multirow{2}{*}{\textbf{Detector}}}                & \multicolumn{2}{c}{\textbf{Features}} & \multicolumn{2}{c}{\textbf{5-Shot}}    & \multicolumn{2}{c}{\textbf{10-Shot}}   & \multicolumn{2}{c}{\textbf{30-Shot}}   \\ \cmidrule(lr){3-4} \cmidrule(lr){5-6} \cmidrule(lr){7-8} \cmidrule(lr){9-10}
\multicolumn{2}{l}{}                                       & \textbf{SPE}          & \textbf{LACS-SP}          & \textbf{mAP}  & \textbf{AP\textsuperscript{50}} & \textbf{mAP}  & \textbf{AP\textsuperscript{50}} & \textbf{mAP}  & \textbf{AP\textsuperscript{50}} \\ \hline
\multirow{7}{*}{YOLOX-S \cite{ge2021yolox}}        & Baseline & & & 5.0 & 10.1 & 9.6 & 18.1 & 14.2 & 26.7                   \\
    & w/ Fang et al. \cite{fang2024data} & & & 5.9  & 11.4 & 11.1 & 20.6 & 15.9 & 27.8 \\
    & w/ InstaGen \cite{feng2024instagen} & & & 11.5  & 19.7 & 15.1 & 25.3 & 18.7 & 30.9 \\
    & w/ InstDiff \cite{wang2024instancediffusion}  & & & 11.0  & 19.2 & 14.9 & 25.6 & 19.5 & 32.6 \\
    & w/ Ours & \checkmark & & 12.4 & 21.4 & 15.6 & 26.8 & 20.1 & 33.5 \\
    & & & \checkmark & 11.4 & 19.7 & 15.7 & 26.5 & 19.7 & 33.0 \\
    & & \checkmark & \checkmark & \textbf{12.8} & \textbf{21.9} & \textbf{16.1} & \textbf{27.1} & \textbf{20.4} & \textbf{34.1} \\ \hline
\multirow{4}{*}{DINO-Swin-L \cite{zhang2022dino}} & Baseline & & & 18.6 & 26.0 & 24.3 & 33.7 & 35.8 & 49.5 \\
    & w/ Fang et al. \cite{fang2024data} & & & 20.3 & 28.1 & 26.0 & 36.8 & 35.0 & 48.8 \\
    & w/ InstaGen \cite{feng2024instagen} & & & 26.4 & 36.0 & 30.4 & 41.3 & 36.4 & 50.0 \\
    & w/ Ours & \checkmark & \checkmark & \textbf{27.4} & \textbf{38.9} & \textbf{30.9} & \textbf{43.8} & \textbf{37.1} & \textbf{53.2} \\ \hline

\label{result:table1}
\end{tabular}
\end{table}

In the standard few-shot settings on COCO \cite{lin2014microsoft}, the 80 dataset categories are split into two sets: 60 base-categories and 20 novel-categories. The base-category images are used to pretrain a detector, while the novel-category images are used to finetune the detector after the initial pretraining. We test our proposed framework on 5-, 10-, and 30-shot settings with YOLOX-S \cite{ge2021yolox} and DINO-Swin-L \cite{zhang2022dino} and report the standard COCO mAP and AP\textsuperscript{50} metrics \cite{everingham2010pascal} on the novel categories. For fair comparison, we use identical hyperparameters for both detector networks as Fang et al. \cite{fang2024data}. 
We observe improvements over both Fang et al. \cite{fang2024data} and state-of-the-art detection dataset generator InstaGen \cite{feng2024instagen} in Table \ref{result:table1} in all scenarios. When utilizing spatial prior extrapolation (SPE) via Mixtral \cite{jiang2024mixtral}, we notice a considerable improvement in mAP due to more diverse box locations, as opposed to simply oversampling the ground truth annotations. When combining SPE with top-1 sample picking via our layout-aware CLIP score, we are able to improve the YOLOX-S \cite{ge2021yolox} baseline by 156\%, 67\%, and 43\% on the 5-, 10-, and 30-shot settings, respectively. Interestingly, running only InstanceDiffusion \cite{wang2024instancediffusion} in the framework brings a noticeable performance boost over the baseline, owing to the benefit of instance-awareness of the LIS model.

\subsubsection{All-Categories Training on Synthetic Data Only}
To test the generalization of our approach compared to InstaGen \cite{feng2024instagen}, we train a standard Faster RCNN \cite{ren2015faster} detector for 12 epochs using synthetic data only from both our proposed framework and InstaGen. Here, we utilize all 80 COCO categories and generate 1284 images using 5-shot layouts as the only available data. In Table \ref{result:table2}, we report additional metrics AP\textsuperscript{75}, mAP for small, medium and large objects (mAP\textsuperscript{s}, mAP\textsuperscript{m}, mAP\textsuperscript{l}), as well as the mean layout-aware CLIP score (mLACS) for both generated datasets. 

\begin{table}[t]
\centering
\caption{Comparison with InstaGen \cite{feng2024instagen} on finetuning Faster RCNN \cite{ren2015faster} using all COCO \cite{lin2014microsoft} categories. We generate 1284 synthetic image-annotation pairs with both our approach and InstaGen \cite{feng2024instagen} and finetune using synthetic data only.}
\begin{tabular}{lccccccc}
\hline
                                       & \textbf{mAP} & \textbf{AP\textsuperscript{50}} & \textbf{AP\textsuperscript{75}} & \textbf{mAP\textsuperscript{s}} & \textbf{mAP\textsuperscript{m}} & \textbf{mAP\textsuperscript{l}} & \textbf{mLACS} \\ \hline
InstaGen \cite{feng2024instagen} & 3.0   & 7.8    & 1.8    & 0.1     & 2.3     & 6.9     & 0.59     \\ \hline
Ours                                   & \textbf{4.3}   & \textbf{10.0}    & \textbf{3.3}    & \textbf{0.7}     & \textbf{5.2}     & \textbf{7.7}     & \textbf{0.65}    
\\ \hline
\label{result:table2}
\end{tabular}
\end{table}

The performance improvements of our method stem from the use of controllable, instance-aware layout-to-image synthesis, in addition to appropriate layout adherence via sample picking.

\subsection{Ablation Studies}
In this section, we ablate over several design choices for our framework. We analyse different LIS models, masks or bounding boxes as conditional layouts, quality vs. quantity in LIS generation, an alternative approach for spatial prior extrapolation, and scalability of our framework with respect to higher augmentation ratios. In all of the ablation studies, we finetune a YOLOX-S \cite{ge2021yolox} in standard COCO \cite{lin2014microsoft} few-shot testing with generated data and evaluate on COCO-val17.

\subsubsection{Comparing LIS Methods and Necessity of Layout-Adherence}
\label{sec:lis_comp}
To generate useful synthetic data for downstream detection tasks, an LIS model boasting both realistic generation capability and layout adherence is of utmost importance. We benchmark three state-of-the-art pretrained LIS models with and without our sample picking: Freestyle \cite{xue2023freestyle}, PLACE \cite{lv2024place}, and InstanceDiffusion \cite{wang2024instancediffusion}. 
In this experiment, we oversample layouts from the 10-shot annotations to reach an augmentation ratio of 4. For generating with mask layouts, we fill bounding boxes with random instance segmentation masks from the 10-shot annotations. 
We assess downstream detection performance along with the layout adherence and quality of the generated instances. To quantify layout adherence and instance quality, we take the mean layout-aware CLIP score (mLACS) over all generated images to measure the former, and calculate the average CLIP \cite{radford2021learning} classification scores on cropped bounding boxes (CS-Crop) for the latter.

\begin{table}[t]
\caption{Ablation over different LIS models \cite{xue2023freestyle, lv2024place, wang2024instancediffusion} and conditioning signals (Cond.), with and without top-1 sample picking w.r.t. our layout-aware CLIP score (LACS-SP) over COCO \cite{lin2014microsoft} 10-shot with YOLOX-S \cite{ge2021yolox}. We record mAP and AP\textsuperscript{50} bounding box metrics as well as dataset quality (Qual.) metrics mean LACS (mLACS) and average CLIP classification score of cropped bounding boxes (CS-Crop). For comparison, we include finetuning results without the use of generated images as baseline.}
\centering
\begin{tabular}{lccccccc}
\hline
\multirow{2}{*}{\textbf{Method}} & \multirow{2}{*}{\textbf{LACS-SP}}& \multicolumn{2}{c}{\textbf{Cond.}} & \multicolumn{2}{c}{\textbf{Box-Metric}}        & \multicolumn{2}{c}{\textbf{Qual.-Metric}} \\ \cmidrule(lr){3-4} \cmidrule(lr){5-6} \cmidrule(lr){7-8}
                                 &    & \textbf{Box}      & \textbf{Mask}      & \textbf{mAP} & \textbf{AP\textsuperscript{50}} & \textbf{mLACS}    & \textbf{CS-Crop}    \\ \hline
Baseline \cite{ge2021yolox}                        &                &                   N/A&                    N/A& 9.6          & 18.1                            & -                 & -                     \\ \hline
Freestyle \cite{xue2023freestyle}                         &                &                   N/A& \checkmark                  & 10.1         & 18.8                            & 0.483             & 0.705                 \\
                                 & \checkmark              &                   N/A& \checkmark                  & 10.7         & 19.6                            & 0.767             & 0.713                 \\ \hline
PLACE \cite{lv2024place}                           &                &                   N/A& \checkmark                  & 10.3         & 19.3                            & 0.447             & 0.774                 \\
                                 & \checkmark              &                   N/A& \checkmark                  & 11.2         & 20.4                            & 0.687             & 0.776                 \\ \hline
Instance-                 &                & \checkmark                 & \checkmark                  & 13.3         & 22.9                            & 0.587 & 0.787 \\
Diffusion \cite{wang2024instancediffusion}                                 & \checkmark & \checkmark                 & \checkmark                  & 13.2         & 22.8                            & \textbf{0.804} & 0.784\\
                                 &                &                   & \checkmark                  & 9.1          & 16.9                            & 0.317 & 0.823 \\
                                 & \checkmark              &                   & \checkmark                  & 9.6          & 17.8                            & 0.563 & \textbf{0.824}\\
                                 &                & \checkmark                 &                    & 14.8         & 25.6                            & 0.522 & 0.817 \\
                                 & \checkmark              & \checkmark                 &                    & \textbf{15.6}         & \textbf{26.7}                            & 0.735 & 0.819\\ \hline

\label{results:table3}
\end{tabular}
\end{table}

As evident from the results from Table \ref{results:table3}, top-1 sample picking with our proposed LACS metric yields a favorable boost in detection performance in almost all cases. This improvement results from discarding noisy images, effectively improving the image-layout alignment in generated images.
All three LIS models benchmarked offer an improvement in downstream detection, with InstanceDiffusion \cite{wang2024instancediffusion} showing the most notable gains. We notice that conditioning InstanceDiffusion with only bounding boxes results in the best improvement, while introducing masks leads to either out-of-layout hallucination (in case of using masks only) or increased instance artifacts (in case of using boxes and masks). One may speculate that the cause could be inaccurate mask pseudo-labelling used in the creation of its training data. 

\subsubsection{Quality vs. Quantity of Generated Images}

\begin{figure}[t]
    \centering
    \begin{subfigure}[b]{0.48\textwidth}
        \centering
        \includegraphics[height=3.6cm, width=5.5cm]{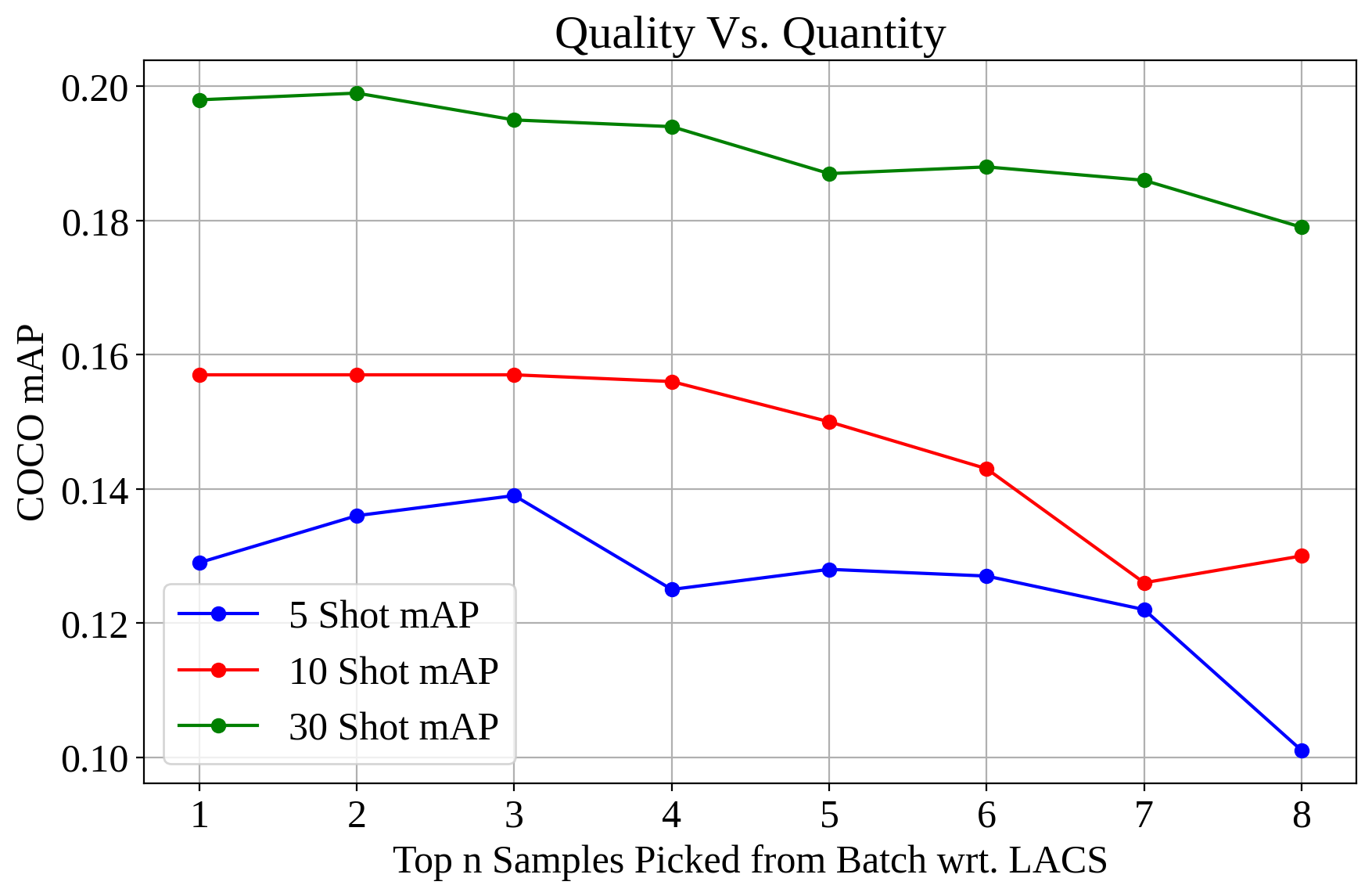}
        \label{fig:qvq_mAP}
    \end{subfigure}
    \hfill
    \begin{subfigure}[b]{0.48\textwidth}
        \centering
        \includegraphics[height=3.6cm, width=5.5cm]{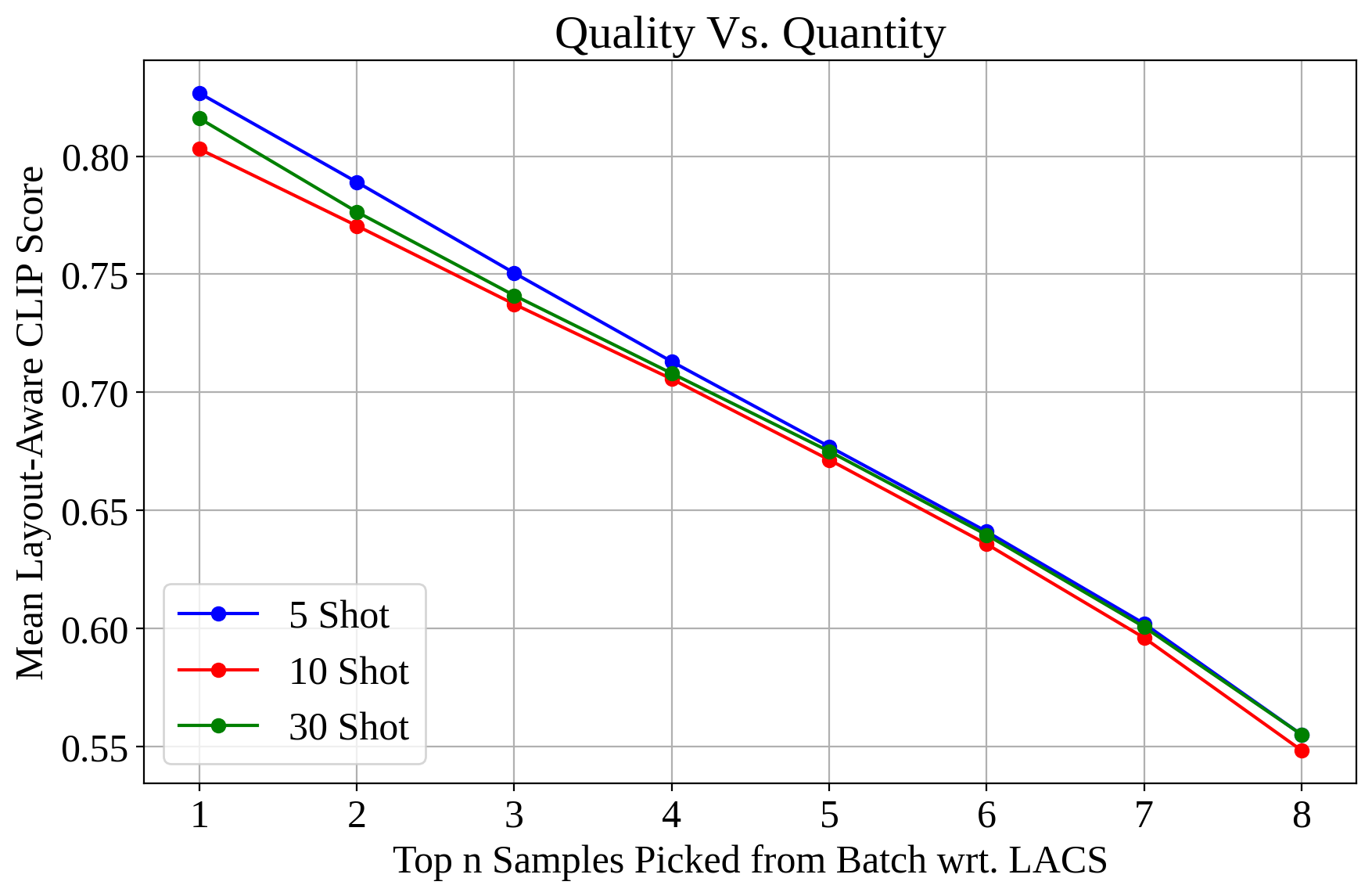}
        \label{fig:qvq_mLACS}
    \end{subfigure}
    \caption{Quality vs. quantity analysis. We analyse the effect of picking the top \textit{n} samples from a generated batch of eight images sorted by the layout-aware clip-score (LACS).}
    \label{fig:qvq}
\end{figure}

One may question whether or not including more samples from the generated batch is beneficial to detection due to scene variance, regardless if the batch contains images not adhering well to the layout. 
To test this, we generate annotations for 5-, 10-, and 30-shot settings with an augmentation ratio of 4 and compare different top-n sample picking strategies: given a conditional layout, we sort a generated batch of eight images according to our LACS metric and pick the top-n scoring samples from the sorted batch. 
To counter-act the effect of longer finetuning, we scale back the number of epochs with respect to the number of samples picked, such that each strategy is trained for the same duration.
Figure \ref{fig:qvq} shows a degradation occurring in detector performance when more than 50\% of samples are picked from the generated batch. This results in overall lower layout-adherence, which is critical for training a detector on LIS synthetic data.

\subsubsection{Alternative Approach for Spatial Prior Extrapolation}

\begin{table}[t]
\caption{Ablation over different spatial prior extrapolation methods. We compare an LLM approach discussed in Section \ref{subsec:spe} with Mixtral \cite{jiang2024mixtral}, ground truth $2\times$ oversampling with x-axis flipping, and N-Ensemble of Gaussian mixture models (GMM). For the GMM approach, we consider drafting object co-occurrences directly from ground truth data (1), or from an auxiliary GMM fitted on the ground-truth object co-occurrence data (2). We also monitor the resulting layout-adherence via mean layout-aware CLIP score (mLACS) discussed in Section \ref{subsec:clipscore}.}
\centering
\begin{tabular}{lccccccc}
\hline
\textbf{Method}               & \textbf{mAP}           & \textbf{AP\textsuperscript{50}}          & \textbf{AP\textsuperscript{75}}          & \textbf{mAP\textsuperscript{s}}        & \textbf{mAP\textsuperscript{m}}         & \textbf{mAP\textsuperscript{l}}         & \textbf{mLACS}          \\ \hline
GT Oversampling + X-flip & 15.6          & 26.7          & 15.9          & \textbf{5.5}          & 14.4          & \textbf{23.2}          & 0.735          \\ \hline
N-Ensemble of GMM (1)    & 14.4          & 25.3          & 14.7          & 4.3          & 13.7 & 21.6          & 0.742          \\
N-Ensemble of GMM (2)    & 14.6          & 25.4          & 15.1          & 4.5          & 12.8          & 22.4          & 0.732          \\ \hline
LLM (Mixtral \cite{jiang2024mixtral})                      & \textbf{15.9} & \textbf{27.0} & \textbf{16.7} & \textbf{5.5} & \textbf{14.9}          & 23.1 & \textbf{0.780}
\\ \hline
\label{results:spe_approaches}
\end{tabular}
\end{table}

\begin{figure}[t]
  \centering
  \includegraphics[width=0.9\textwidth]{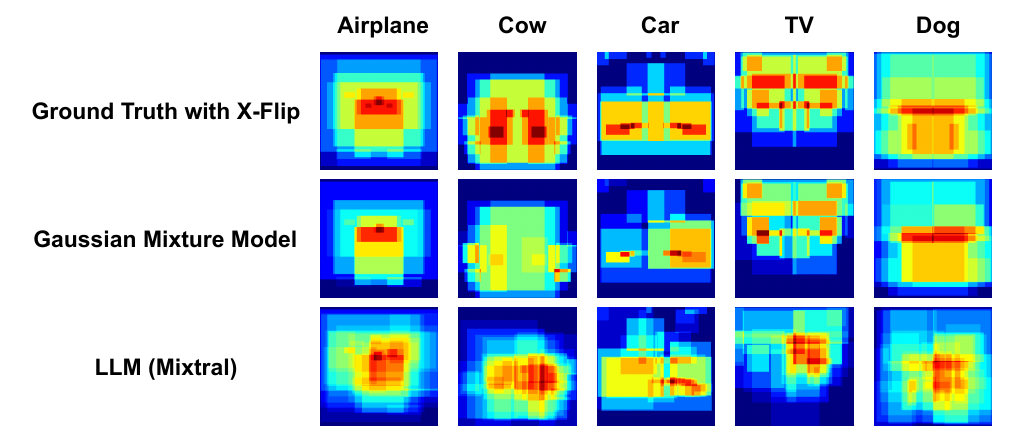}
  \caption{Bounding box heatmaps for ground truth boxes, Gaussian mixture model generated boxes, and Mixtral \cite{jiang2024mixtral} generated boxes on 5 novel COCO \cite{lin2014microsoft} categories.
  }
  \label{fig:bbox_heatmap}
\end{figure}
In this ablation study, we explore an alternative approach to spatial prior extrapolation with Gaussian Mixture Models (GMM). Specifically, we model bounding boxes of shape \(bbox \in \mathbb{R}^{x \times y \times w \times h}\) as a random distribution and fit an ensemble of \textbf{\textit{N}} GMMs on the bounding boxes of novel categories (one category per GMM). After fitting, we sample valid bounding boxes from the fitted ensemble of GMMs according to object co-occurrences from a ground truth layout (e.g. how many objects of category 'person', 'car', 'cat', etc). We also explore fitting an additional GMM to model object co-occurrences as a random distribution as well.
We benchmark both GMM approaches (with and without object co-occurrence modelling) by fitting them on COCO \cite{lin2014microsoft} 10-shot annotations and generating bounding box layouts with the default augmentation ratio $\alpha$ of 4. We compare this approach against using Mixtral \cite{jiang2024mixtral} for layout generation as outlined in Section \ref{subsec:spe} and a simple strategy of oversampling the ground truth annotations with x-axis flipping. 
Results shown in Table \ref{results:spe_approaches} indicate that modelling layouts as random distributions via GMMs does not yield improvements over ground truth oversampling, which can be attributed to layout data scarcity in few-shot settings. Interestingly, Mixtral \cite{jiang2024mixtral} is able to extrapolate reasonably from the provided examples and produces plausible and diverse bounding box layouts. Figure \ref{fig:bbox_heatmap} shows the bounding box heatmaps for ground truth boxes, GMM-sampled boxes, and Mixtral-sampled boxes.

\subsubsection{Effect of Higher Augmentation Ratios}
\label{sec:aug_ratios}

\begin{figure}[t]
  \centering
  \includegraphics[width=\textwidth]{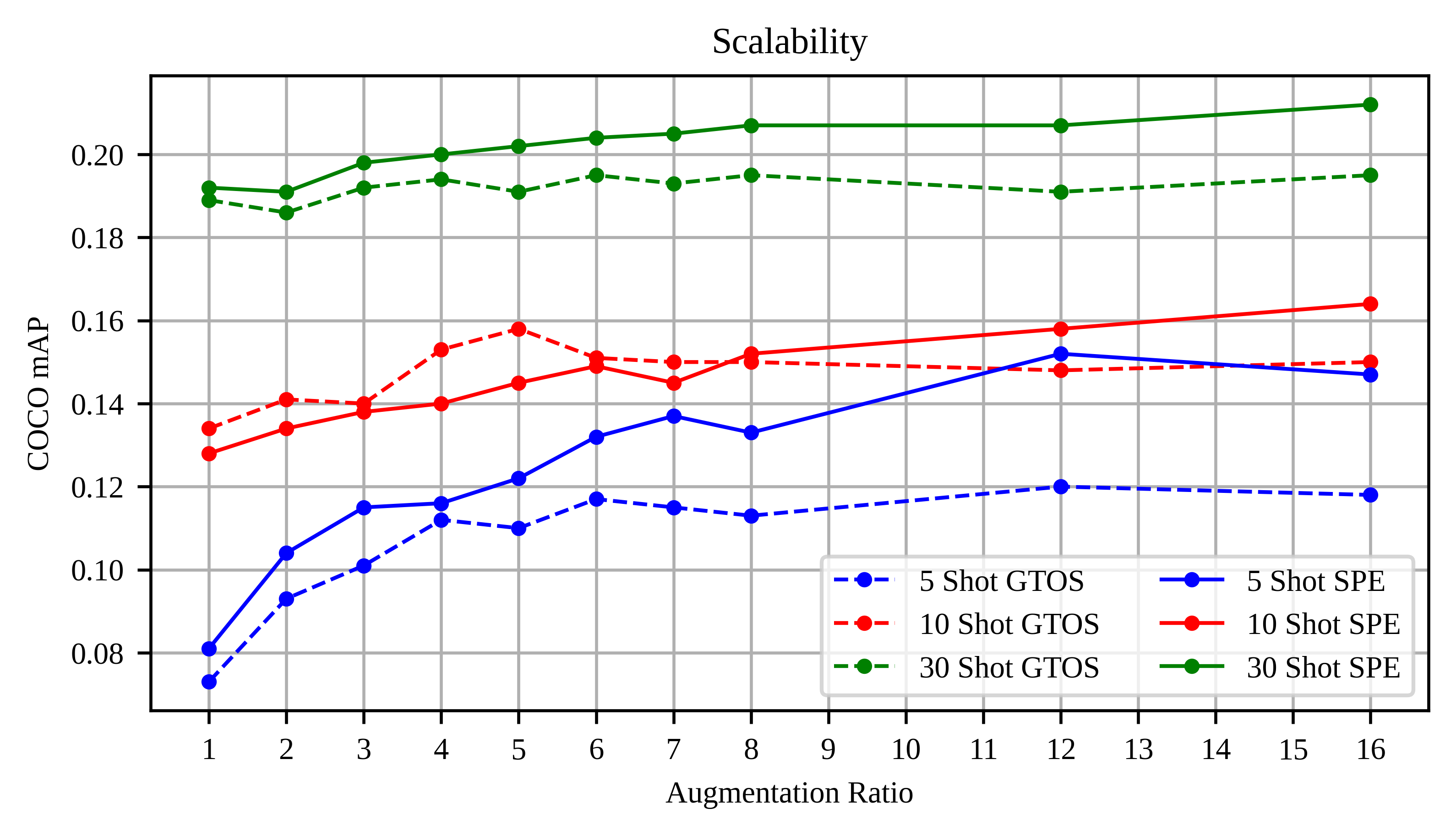}
  \caption{Scalability of our method with respect to higher augmentation ratios $\alpha$. We generate bounding box annotations with our LLM-based spatial prior extrapolation (SPE) method and compare with simple ground truth oversampling with x-axis flipping (GTOS). We synthesize images conditioned on the generated box layouts with InstanceDiffusion \cite{wang2024instancediffusion} and subsequently evaluate on YOLOX-S \cite{ge2021yolox} finetuning.
  }
  \label{fig:scalability}
\end{figure}

To showcase the full extent of our generative augmentation framework, we incrementally increase the layout augmentation ratio $\alpha$ and monitor the downstream YOLOX-S \cite{ge2021yolox} performance. We compare ground truth layout oversampling (GTOS) against our LLM-based spatial prior extrapolation (SPE). Figure \ref{fig:scalability} shows superior scalability when generating with SPE than with GTOS. This effect is more pronounced on augmentation ratios of 8 and higher. In 10-shot settings, SPE lags slightly behind GTOS, but begins to outperform GTOS on higher augmentation ratios. In 5-shot and 10-shot settings, SPE remains superior across all augmentation ratios.

\section{Conclusion}
\label{sec:conc}

We propose a generative augmentation framework for few-shot object detection, combining a large language model with a layout-to-image synthesis model to generate high-quality image-annotation pairs. We introduce the Layout-Aware CLIP Score (LACS), a CLIP-based \cite{radford2021learning} metric for evaluating layout adherence in generated images. Our framework outperforms state-of-the-art generative methods \cite{fang2024data, feng2024instagen} on few-shot COCO \cite{lin2014microsoft} benchmarks, and ablation studies demonstrate the benefits of LLM-based spatial prior extrapolation and improving layout adherence through sample selection using LACS.

\subsubsection{\ackname} We would like to thank Haoyang Fang \cite{fang2024data} for their help in reproducing their experiment baselines.

%
%
\bibliographystyle{splncs04}
\bibliography{main}
\end{document}